\documentclass[letterpaper]{article} 
\usepackage{aaai25}  
\usepackage{times}  
\usepackage{helvet}  
\usepackage{courier}  
\usepackage[hyphens]{url}  
\usepackage{graphicx} 
\usepackage{natbib}  
\usepackage{caption} 
\usepackage{amsmath,amssymb,amsfonts}
\usepackage[ruled,linesnumbered]{algorithm2e}
\usepackage{textcomp}
\usepackage{xcolor}
\usepackage{float}
\usepackage{booktabs}
\usepackage{multirow}
\usepackage {subcaption}
\usepackage{float}
\usepackage{makecell}

\usepackage{newfloat}
\usepackage{listings}
\DeclareCaptionStyle{ruled}{labelfont=normalfont,labelsep=colon,strut=off} 
\floatstyle{ruled}
\newfloat{listing}{tb}{lst}{}
\floatname{listing}{Listing}
\title{Towards Scalable and Deep Graph Neural Networks via Noise Masking}

\author {
    Yuxuan Liang\textsuperscript{\rm 1},
    Wentao Zhang\textsuperscript{\rm 2}\footnote{Corresponding author},
    Zeang Sheng\textsuperscript{\rm 3},
    Ling Yang\textsuperscript{\rm 3},
    Quanqing Xu\textsuperscript{\rm 4},\\
    Jiawei Jiang\textsuperscript{\rm 5},
    Yunhai Tong\textsuperscript{\rm 1},
    Bin Cui \textsuperscript{\rm 3,6}\footnotemark[1]
}

\affiliations {
    \textsuperscript{\rm 1}School of Intelligence Science and Technology, Peking University\\
    \textsuperscript{\rm 2}Center for Machine Learning Research, Peking University\\
    \textsuperscript{\rm 3}School of CS \& Key Laboratory of High Confidence Software Technologies (MOE), Peking University\\
    \textsuperscript{\rm 4}OceanBase, Ant Group
    \textsuperscript{\rm 5}School of Computer Science, Wuhan University\\
    \textsuperscript{\rm 6}Institute of Computational Social Science, Peking University (Qingdao)\\
    \{liangyx, yangling\}@stu.pku.edu.cn, \{wentao.zhang, shengzeang18, yhtong, bin.cui\}@pku.edu.cn, \\
    jiawei.jiang@whu.edu.cn,
    xuquanqing.xqq@oceanbase.com
}
 
\usepackage{bibentry}

\begin{document}

\maketitle

\begin{abstract}
In recent years, Graph Neural Networks (GNNs) have achieved remarkable success in many graph mining tasks.
However, scaling them to large graphs is challenging due to the high computational and storage costs of repeated feature propagation and non-linear transformation during training.
One commonly employed approach to address this challenge is model-simplification, which only executes the \textbf{P}ropagation (\textbf{P}) once in the pre-processing, and  
\textbf{C}ombine (\textbf{C}) these receptive fields in different ways and then feed them into a simple model for better performance.
Despite their high predictive performance and scalability, these methods still face two limitations.
First, existing approaches mainly focus on exploring different \textbf{C} methods from the model perspective, neglecting the crucial problem of performance degradation with increasing \textbf{P} depth from the data-centric perspective, known as the over-smoothing problem.
Second, pre-processing overhead takes up most of the end-to-end processing time, especially for large-scale graphs.
To address these limitations, we present random walk with noise masking (RMask), a plug-and-play module compatible with the existing model-simplification works. 
This module enables the exploration of deeper GNNs while preserving their scalability.
Unlike the previous model-simplification works, we focus on continuous \textbf{P} and found that the noise existing inside each \textbf{P} is the cause of the over-smoothing issue, and use the efficient masking mechanism to eliminate them.
Experimental results on six real-world datasets demonstrate that model-simplification works equipped with RMask yield superior performance compared to their original version and can make a good trade-off between accuracy and efficiency.
\end{abstract}




\section{Introduction}
Graph Neural Networks (GNNs) have achieved great success in graph representation learning. In recent years, GNNs have been widely used in many graph-based applications, such as databases~\cite{database1, database2, database3}, data management~\cite{datamanagement1, dse1, dse2}, and data mining~\cite{scis1, scis2, jcst1, jcst2}.
Despite the success of GNNs, scaling them to large graphs is challenging due to the high computational and storage costs of repeated feature propagation during training.
Consequently, these limitations impede the scalability of GNNs on large graphs, limiting their applicability and advancement in real-world scenarios.

\begin{figure*}[t]
    \centering
    \begin{subfigure}{0.69\linewidth}
        \includegraphics[width=0.55\linewidth]{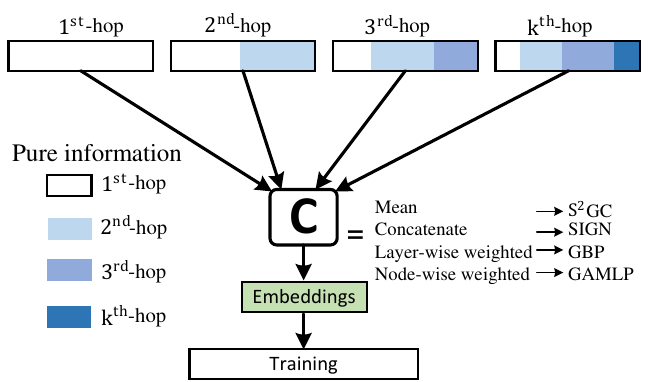}
        \includegraphics[width=0.4\linewidth]{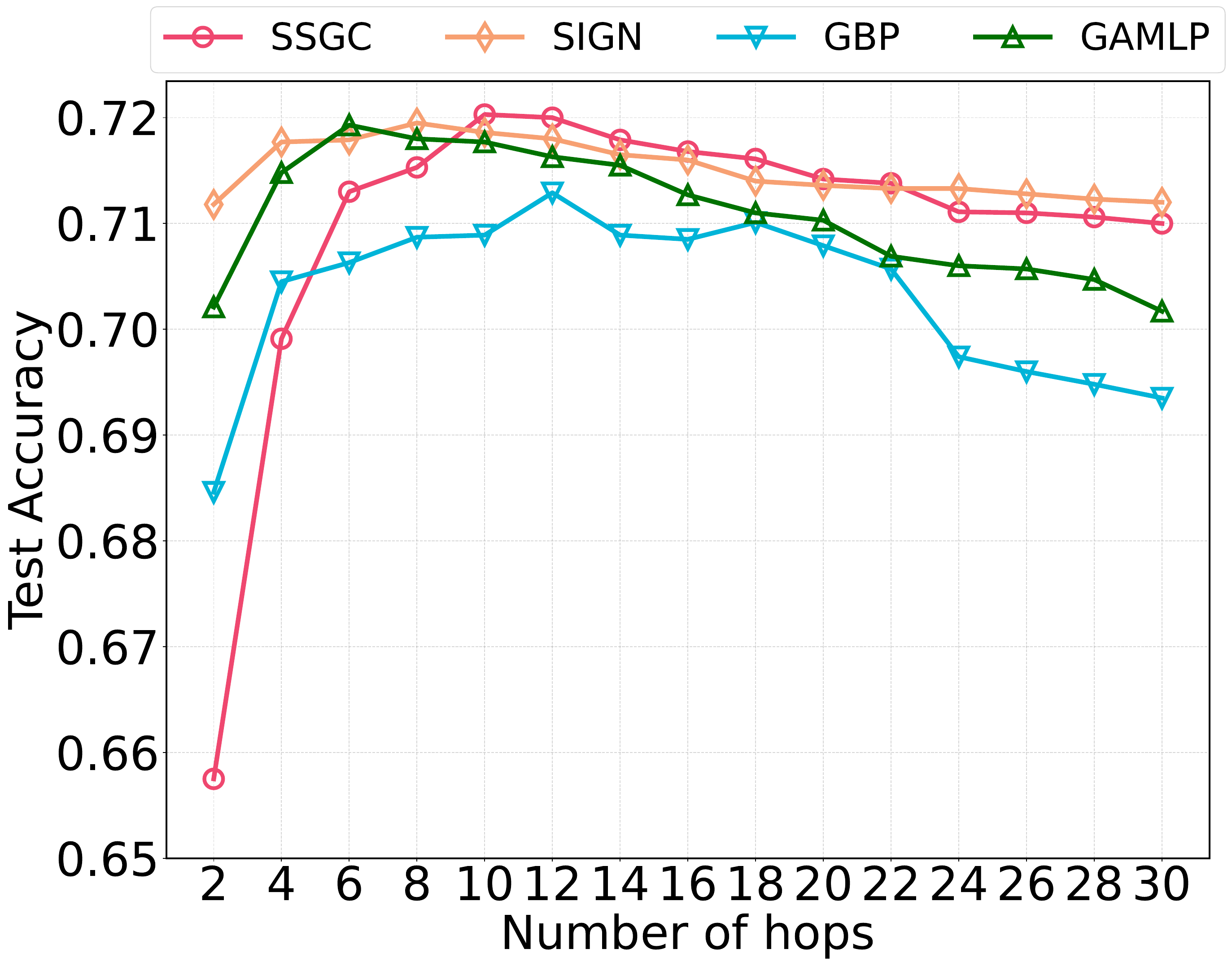}
        \caption{Propagation with Noise Information: (Left) Illustration of existing model-simplification GNNs. (Right) Illustration of over-smoothing problem.}
    \end{subfigure}
    \begin{subfigure}{0.29\linewidth}
        \includegraphics[width=1.0\linewidth]{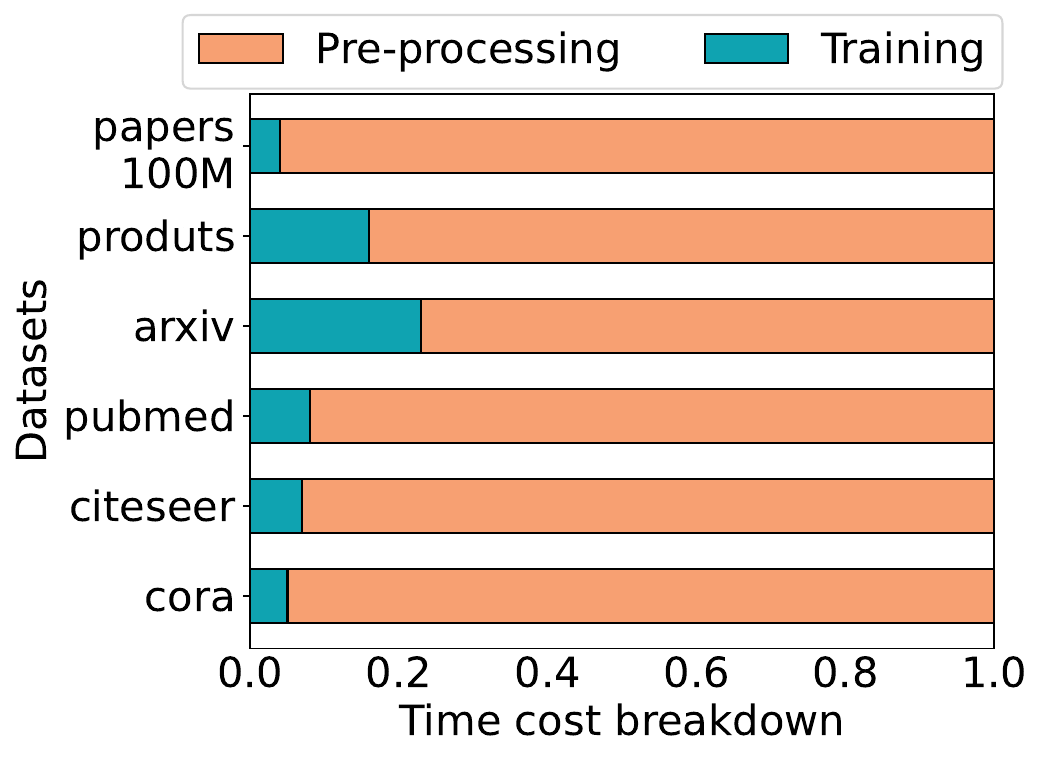}
        \caption{Propagation with high pre-processing overhead.}
    \end{subfigure}
    \caption{Example of two limitations in model-simplification GNNs.}
\label{fig:1}
\end{figure*}

To address the scalability issues,  \textit{model-simplification GNNs}, as a promising direction for scalable performance has aroused great interest recently.
The most representative work is SGC~\cite{SGC}, which puts feature propagation in the pre-processing step by removing nonlinearities and collapsing weight matrices between consecutive layers.
Following the design principle of SGC, piles of works have been proposed to improve the performance of SGC further while maintaining high scalability, such as SIGN, S$^2$GC, GBP, GAMLP~\cite{SIGN, S2GC, GBP, GAMLP}.
Generally, these model-simplification GNNs can be disentangled into two independent operations:
\emph{Propagation} (\textbf{P}) and \emph{Combination} (\textbf{C}).
The \textbf{P} operation can be viewed as aggregating the information of first-order neighbors for each node.
The \textbf{C} operation can be viewed as a combination of consecutive \textbf{P} operations. Continuous \textbf{P} operations are used to capture deeper receptive fields and \textbf{C} operations are used to combine these receptive fields for better performance.
Existing model-simplification GNNs mainly focus on designing different \textbf{C}.
Despite their high scalability and predictive performance, existing model-simplification GNNs still face the following two limitations:

\textbf{Propagation with Noise Information.}
Despite the success of model-simplification GNNs, the exploration of deep propagation steps remains limited because simply stacking \textbf{P} leads node representations to become indistinguishable and results in performance degradation, i.e., over-smoothing problem.
Existing model-simplification GNNs, as shown in Figure~\ref{fig:1}(a) (left), pay more attention to designing different combination methods of propagation steps without considering the noise problem within each hop.
 From Figure~\ref{fig:1}(a) (left), we can see that each hop contains overlapping graph structure information of the previous hop, such noise information will be propagated along the edges and hinder the extraction of truly useful high-hop information.
As shown in Figure~\ref{fig:1}(a) (right), we conducted an experiment using four model-simplification GNNs on the ogbn-arxiv dataset. Experimental results show that the performance continues to decline as the propagation depth increases.

\textbf{Propagation with High Pre-processing Overhead.}
Even though existing model-simplification GNNs are significantly faster than traditional GNNs~\cite{GCN, GAT} by putting the expensive feature propagation step into the pre-processing step and performing it only once, pre-processing overhead still accounts for a large proportion of the entire training time.
As shown in Figure~\ref{fig:1}(b), we performed end-to-end training time cost breakdown using the SGC model on Cora, Citeseer, Pubmed, ogbn-arxiv, ogbn-products and ogbn-papers100M datasets~\cite{GCN,ogb}.
We can see that the pre-processing time occupies the vast majority of the overall training time. Taking ogbn-papers100M as an example, the end-to-end training time takes 4.4 hours, and pre-processing overhead accounts for 96\%.


In this paper, we introduce random walk with noise masking (RMask), a plug-and-play module that seamlessly integrates with existing model-simplification GNNs.
By tracking the information of each hop in the pre-processing step, we found that the continuous $\textbf{P}$ operations generate a significant amount of noise information, hindering the ability towards deeper GNNs.
Based on this observation, our key insight is to employ a noise masking mechanism to extract valuable high-hop information and mitigate the over-smoothing issue.
Furthermore, to tackle the issue of high pre-processing cost in model-simplification GNNs, we utilize the de-noise-based random walk method to extract pure graph structure information. This approach enables us to strike a balance between accuracy and efficiency effectively.

This paper does not intend to diminish the contribution of current methods for eliminating over-smoothing, like DropEdge~\cite{DropEdge}, GPR~\cite{GPR}, DAGNN~\cite{DAGNN}.
Instead, we aim to provide new insights into the over-smoothing problem from the $\textbf{P}$ operation itself in scalable GNNs, which is orthogonal to other methods to eliminate over-smoothing.

\textbf{Contirbutions.}
The main contributions of this work are as follows:
\textit{\underline{New findings}}. In contrast to previous model-simplification GNNs that primarily focused on studying the combination methods ($\textbf{C}$) between different hops, our research delves into exploring the noise information within each hop. We discover that the $\textbf{P}$ operations introduce a significant amount of noise, exacerbating the over-smoothing problem.
\textit{\underline{New method}}. We introduce RMask, a plug-and-play module compatible with existing model-simplification GNNs. To address the noise issue caused by $\textbf{P}$ operations, we propose a noise masking mechanism that extracts pure information within each hop. Additionally, we leverage random walks to strike a balance between accuracy and efficiency.
\textit{\underline{State-of-the-art performance}}. We evaluate the effectiveness of RMask on three widely used datasets (Cora, Citeseer, Pubmed)~\cite{GCN} and three large-scale datasets (ogbn-arxiv, ogbn-products, ogbn-papers100M)~\cite{ogb}. Experimental results show that model-simplification GNN equipped with RMask has superior performance compared to itself.

\begin{figure*}[t]
    \centering
    \begin{subfigure}{0.32\linewidth}
        \includegraphics[width=\linewidth]{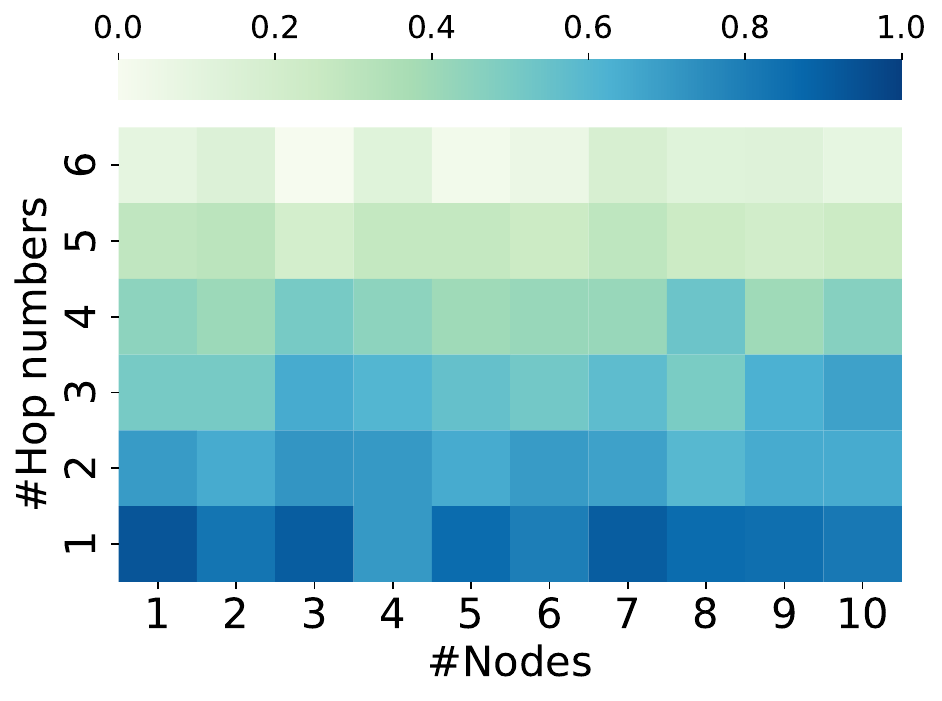}
        \caption{Weight distribution with different hops.}
    \end{subfigure}
    \begin{subfigure}{0.32\linewidth}
        \includegraphics[width=\linewidth]{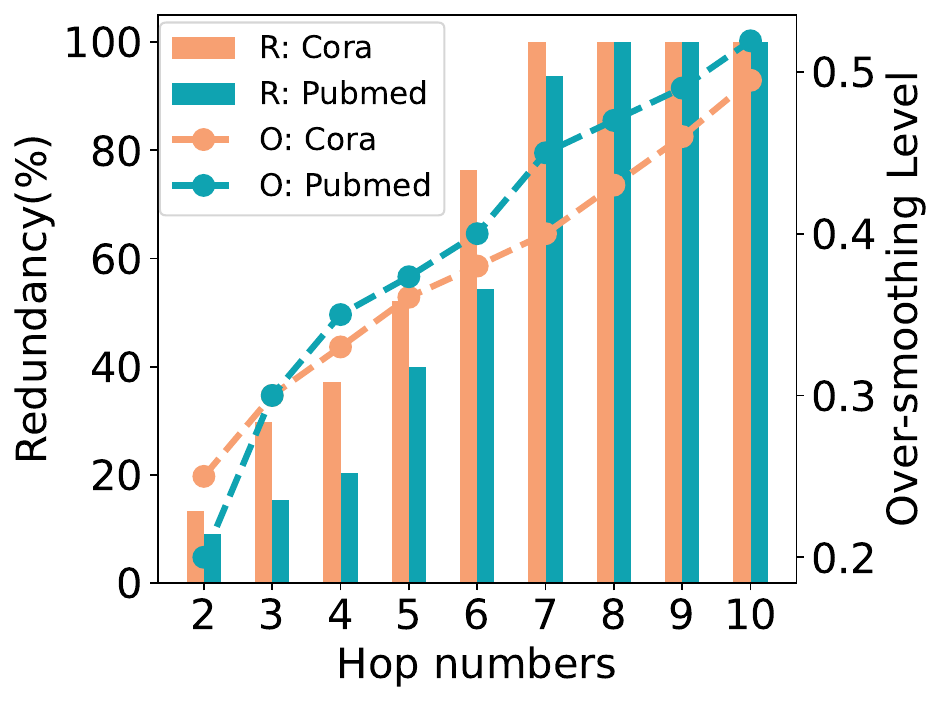}
        \caption{Smoothness level and noise information.}
    \end{subfigure}
    \hspace{0.2cm}
    \begin{subfigure}{0.32\linewidth}
        \includegraphics[width=0.82\linewidth]{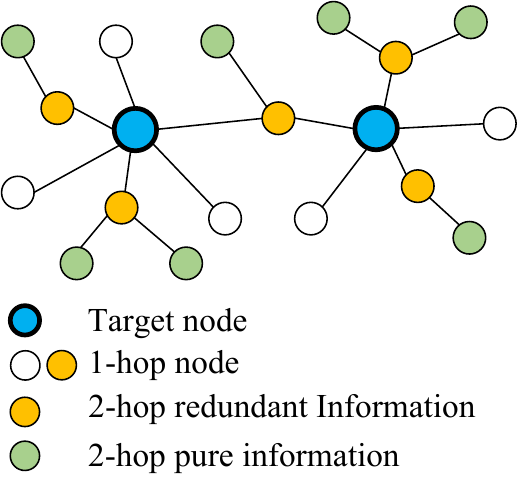}
        \caption{Redundant information.}
    \end{subfigure}
    \begin{subfigure}{0.32\linewidth}
        \includegraphics[width=\linewidth]{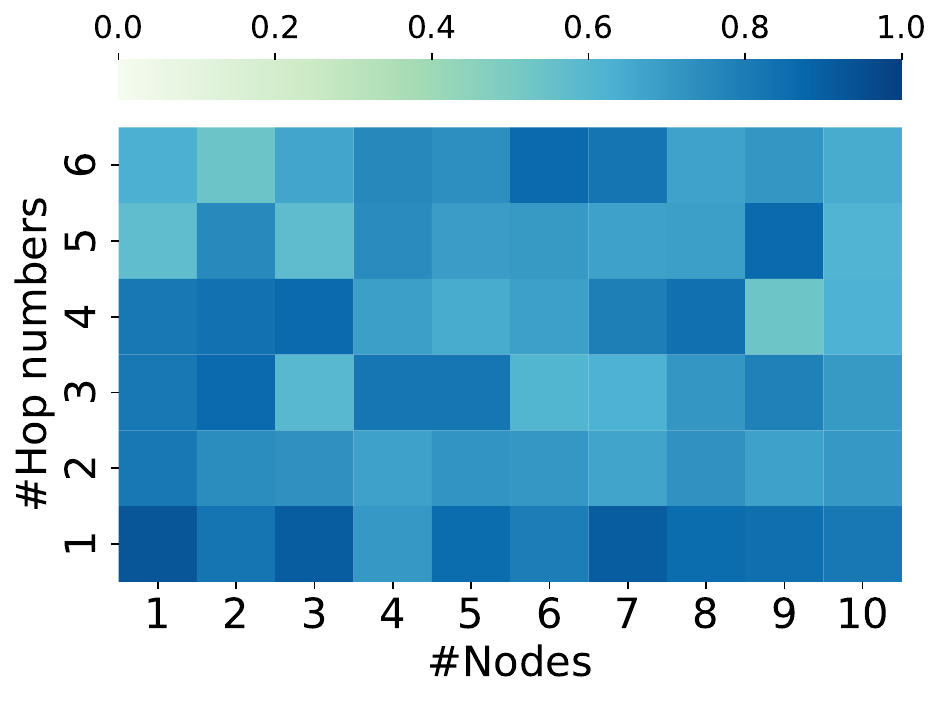}
        \caption{Weight distribution with noise masking.}
    \end{subfigure}
    \begin{subfigure}{0.32\linewidth}
        \includegraphics[width=\linewidth]{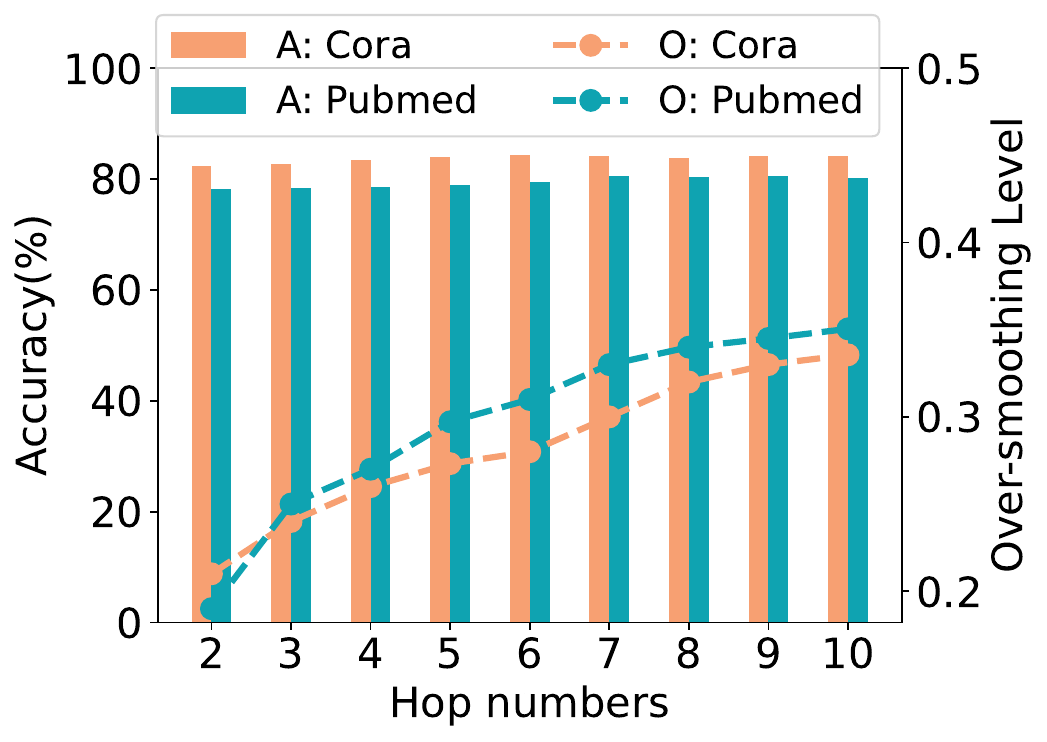}
        \caption{Smoothness level with noise masking.}
    \end{subfigure}
    \begin{subfigure}{0.32\linewidth}
        \includegraphics[width=\linewidth]{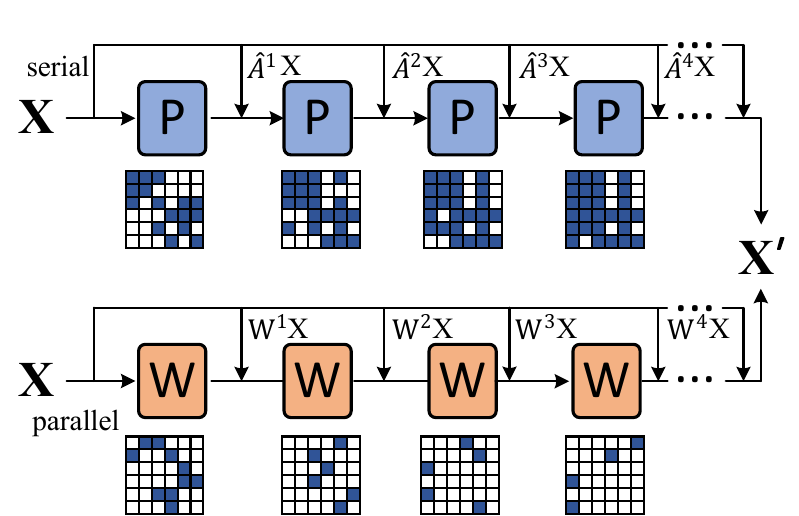}
        \caption{High pre-processing overhead.}
    \end{subfigure}
    \caption{Observation and our insight.}
\label{fig:3}
\end{figure*}

\section{Preliminaries}
\textbf{Notations and Problem Formulation.}
We consider an undirected graph $\mathcal{G}=(\mathcal{V}, \mathcal{E})$ comprising a set of nodes $\mathcal{V} = {v_1, \cdots, v_n}$ and a set of edges $\mathcal{E} = \{(v_i, v_j) | v_i, v_j \in \mathcal{V}\}$. The cardinality of the node and edge sets is $|\mathcal{V}| = N$ and $|\mathcal{E}| = M$, respectively.
$\mathbf{A} \in \mathbb{R}^{|N| \times |N|}$ represents the adjacency matrix of $\mathcal{G}$. Besides, we define the degree matrix of $\mathbf{A}$ as $\mathbf{D}$, which is a diagonal matrix with entries corresponding to the degrees of the nodes. Specifically, $\mathbf{D} = diag(d_1, \cdots, d_n)\in \mathbb{R}^{N \times N}$, where $d_i =\sum_{j \in \mathcal{V}} \mathbf{A}_{ij}$ represents the degree of node $i$.
Each node has a feature vector $\in \mathbb{R}^d$, resulting in an $N \times d$ matrix $\mathbf{X}$ when stacked together. Here, $\mathbf{X}_i$ refers to the feature vector of node $i$.

\textbf{Smoothness Level.}
To measure the over-smoothing levels, we introduce the concept of {Smoothness Level} in DAGNN~\cite{DAGNN} and employ cosine similarity to measure the similarity between two nodes. 
 We formally define the ``Node Smoothness Level (NSL)'' and ``Graph Smoothness Level (GSL)'' as follows: $NSL_{i} = \frac{1}{N-1}\sum_{j \in \mathcal{V}, j \neq i}\frac{\mathbf{X}_i \cdot \mathbf{X}_j}{|\mathbf{X}_i||\mathbf{X}_j|}$.
$GSL = \frac{1}{N}\sum_{j \in \mathcal{V}}NSL_i$.
$NSL_i$ computes the average similarity between node $i$ and all other nodes in the graph, while $GSL$ calculates the average similarity between pairs of nodes in the graph. A higher smoothness level indicates a higher probability of similarity between two randomly selected nodes from a given set.
In this paper, we use GSL to measure the smoothness level.

\section{Related Work}
\textbf{Sampling Graph Neural Networks.}
Over the past few years, the graph convolution operation introduced by GCN has increasingly become the standard form in most GNN architectures~\cite{GCN}.
GCN adopts a layer-wise propagation rule and a multi-layer non-linear feature transformation network to form the new representation.
However, this approach has expensive message propagation overhead during training, resulting in computationally expensive and low scalability.
A commonly used method to tackle the scalability issue is sampling, which focuses on a smaller portion of the graph while still preserving its structural properties. 
Existing methods typically employ sampling techniques at various levels: node-level samplings, such as GraphSAGE~\cite{GraphSAGE} and VR-GCN~\cite{VR-GCN}; layer-level samplings, such as Fast-GCN~\cite{FastGCN} and ASGCN~\cite{ASGCN}; and graph-level samplings, such as ClusterGCN~\cite{Cluster-GCN} and GraphSAINT~\cite{GraphSAINT}.


\textbf{Model-simplification Graph Neural Networks.}
The other direction is to build model-simplification GNNs.
The main idea is to decouple the feature propagation and non-linear transformation in the GNN layer and finish the time-consuming feature propagation process without model parameter training.
SGC~\cite{SGC} removes nonlinearities between consecutive graph convolutional layers, leading to higher scalability and efficiency.
Specifically, SGC performs \textbf{P} operations as follows:
\begin{equation}
\begin{aligned}
&\mathbf{X}^{(k)} = (\tilde{\mathbf{D}}^{r-1}\tilde{\mathbf{A}}\tilde{\mathbf{D}}^{-r})^k\mathbf{X}^{(0)}
\end{aligned}
\label{SGC}
\end{equation}
Where $\mathbf{X}^{(0)}$ represents the raw feature, and $\mathbf{\hat{A}} = \mathbf{\tilde{D}}^{r-1}\mathbf{\tilde{A}}\mathbf{\tilde{D}}^{-r}$ represents the normalized adjacency matrix $\mathbf{\tilde A}$, where $\mathbf{\tilde{D}}$ is the degree matrix of $\mathbf{\tilde A}$.  
Following the design principle of SGC, piles of works have been proposed to further improve the performance of SGC while maintaining high scalability and efficiency, such as SIGN~\cite{SIGN} S2GC~\cite{S2GC}, GBP~\cite{GBP} and GAMLP~\cite{GAMLP}.

\textbf{Over-smoothing Problem.}
By taking a large \textbf{P} step, GNNs allow each node to capture deeper graph structure information.
However, an excessively large propagation step can result in indistinguishable node embeddings, despite the advantages it offers.
If we apply the $\textbf{P}$ for infinite times, the node representations within the same connected component would reach a stationary state~\cite{NAFS}, leading to indistinguishable outputs. Concretely, when adopting
$\mathbf{\hat{A}} = \mathbf{\tilde{D}}^{r-1}\mathbf{\tilde{A}}\mathbf{\tilde{D}}^{-r}$, $\mathbf{\hat{A}^{\infty}}$ follows
$\mathbf{\hat{A}}^{\infty}_{i,j} = \frac{(d_i+1)^r(d_j+1)^{1-r}}{2M+N}, ~\mathbf{X}^{\infty} = \mathbf{\hat{A}}^{\infty}\mathbf{X}^{0}$
which shows that as the depth of \textbf{P} approaches $\infty$, the influence from node $j$ to node $i$ is only determined by their node degrees.

Previous research usually addresses the over-smoothing problem from three aspects: manipulating graph topology~\cite{DropEdge}
, refining model structure~\cite{GPR}, and dynamic learning~\cite{DAGNN}.
However, there is no method to consider over-smoothing from the perspective of noise information during feature propagation.

\begin{figure*}[t]
\centerline{\includegraphics[width=6in]{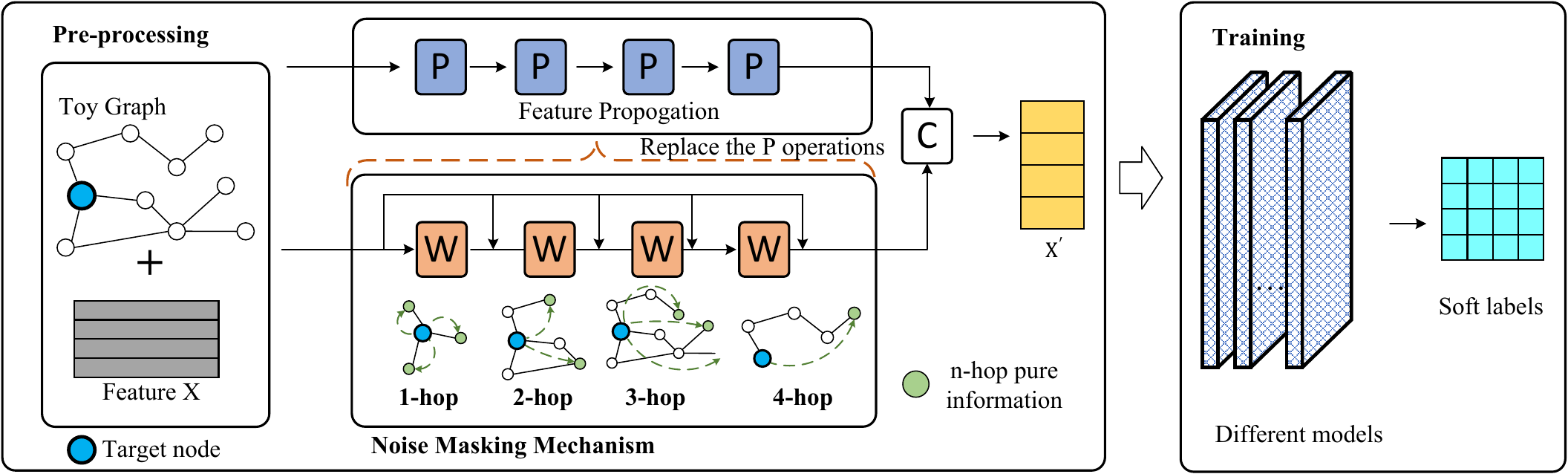}}   
\caption{The architecture of proposed RMask.}
\label{fig:framework}
\end{figure*}

\section{Motivatoin}
In this section, we make a deep analysis of the two limitations that exist in model-simplification GNNs and then provide our insight to help us design the architecture of RMask.

\textbf{Study on Noise Information.}
\label{sec 3.1}
Model-simplification GNNs implement $\textbf{P}$ operations by continuously performing matrix product operations on the initial normalized matrix $\hat A$ using Eq. (1) to increase the propagation depth (i.e., the number of hops).
However, this approach will weaken the importance of high-hop information.
We randomly select 10 nodes on the Cora dataset and observe the average weight of each hop through \textbf{P} operations with L2 normalization.
As shown in Figure~\ref{fig:3}(a), nodes with higher weights are frequently captured within lower hops,  while nodes holding valuable information in higher hops exhibit considerably lower weights. 
This phenomenon hinders the capture of higher-hop information.
To explain further, we conduct a 2-hop propagation starting from the target node.
As shown in Figure~\ref{fig:3}(c), the information captured by 2-hop encompasses not only the current hop but also 2-hop redundant information, since this information can already be captured within 1-hop, we refer to it as \emph{Noise Information}.

As the propagation depth increases, the nodes captured by high hop contain a large amount of low hop noise information, making it difficult to distinguish between high hop and low hop information, exacerbating the over-smoothing problem.
To further examine the impact of noise information on over-smoothing, we increase the hop numbers and measure the proportion of noise information and GSL using the SIGN model.
As shown in Figure~\ref{fig:3}(b), GSL grows explosively with the increase of hop number, and the noise information also grows continuously. The information captured after 7 hops is completely redundant. 

\emph{Insight 1: Noise information will hinder the utilization of high-hop information and aggravate over-smoothing.}
We re-implemented SIGN with noise masking. As shown in Figure~\ref{fig:3}(d), the node can not only capture the effective information of higher hop but also eliminate the over-smoothing problem as shown in Figure~\ref{fig:3}((e). As the number of hops increases, the accuracy and smoothness level tends to be flat.


\textbf{Study on High Pre-processing Overhead.}
\label{sec 3.2}
Moreover, this propagation method results in significant pre-processing overhead. The upper part of Figure~\ref{fig:3}(f) illustrates the unified pre-processing process employed by current model-simplification GNNs.
First, the time complexity of preprocessing is linearly related to the number of edges~\cite{GBP}. Each hop captures a significant amount of graph structural information from all previous hops, which leads to computational intensity due to the dense matrix involved.
Second, this approach relies on the interdependence of information across different hops, which can only be obtained serially.
Compared with expensive pre-processing overhead, model-simplification CNNs usually use simple models for fast training.
For the above reasons, the pre-processing overhead constitutes the majority of the end-to-end training time, as demonstrated in Figure~\ref{fig:1}(b).

\emph{Insight 2:
As shown in the lower part of Figure~\ref{fig:3} (f), to reduce the high overhead of pre-processing, we need a sparse and parallel method to efficiently capture the important information of each hop.}

\textbf{Proposal of RMask.}
Motivated by the two insights, on one hand, we can eliminate noise information by identifying redundant information of each hop and using the masking mechanism. On the other hand, we can use random walks to capture truly useful information for different hops in a highly parallel manner. Additionally, the masking mechanism produces a sparse graph, further reducing the computational overhead of aggregation.

\section{Method}
In this section, we introduce RMask, a plug-and-play module designed for model-simplification GNNs as illustrated in Figure~\ref{fig:framework}. 
The noise masking mechanism consists of two main components: noise information identification and neighbor nodes importance assignment. 
In this section, we will explain the pipeline of RMask and its methods in detail. In addition, we also analyze the time complexity of RMask.

\subsection{RMask Pipeline}
Existing model-simplification GNNs follow a common pipeline consisting of two main steps (upper part of Figure~\ref{fig:framework}). They employ the same $\textbf{P}$ operation and emphasize the combination (\textbf{C}) of information from different hops. 
In contrast, 
RMask utilizes a random walk noise masking mechanism (\textbf{W} operations) to replace the \textbf{P} operations, aiming to address the over-smoothing issue and extract the truly useful information from high-hop (lower part of Figure~\ref{fig:framework}).
Given a specified number of hops and a graph structure, we first perform random walks with the masking mechanism for each node based on the graph structure. Then the captured graph structure information and features are aggregated to obtain results for different hops.
Furthermore, the feature propagation results obtained in this manner can directly replace \textbf{P} operations in other model-simplification GNNs (such as S$^2$GC, GBP, SIGN, GAMLP, etc.). Simultaneously, we retain the advantages of feature combination and model selection from existing model-simplification GNNs.

\begin{algorithm}[t]
\label{algorithm}
\caption{The overall process of RMask}\label{algorithm}
\SetKwData{Left}{left}\SetKwData{This}{this}\SetKwData{Up}{up}
\SetKwInOut{Input}{Input}\SetKwInOut{Output}{Output}
\Input{Graph $\mathcal{G}$, hops for propogation $H$, number for random walks $T$}
\Output{De-noise random walk matrices $\mathbf{W}$}
$\mathbf{M} = \bigcup_{h=1}^{H}\mathbf{M}^h$ with Eq. (2)\\
$\mathbf{S} = \alpha(I-(1-\alpha)\hat{A})^{-1}$ with Eq. (4)\\
\For{$h \in H$}{
    \For{$t \in T$}{
        $\mathbf{W}^h_t = RW(\mathcal{G}, M^h, \mathbf{S})$ with Eq. (5)\\
    }
    $\mathbf{W}^h = \frac{\bigcup_{t \in T}W^h_t}{T}$\\
}
$\mathbf{W} = concatenate([\mathbf{W}^{h1}||\mathbf{[W}^{h2}||...])$\\
\textbf{Return:} {$\mathbf{W}$}
\end{algorithm}

\subsection{Noise Masking Mechanism}
The key insight of the noise masking mechanism is that if the information captured at a higher hop is already encompassed by the information captured at lower hops, then this noise information needs to be masked in the higher hop.
Based on this insight, our noise masking mechanism consists of two components: noise information identification and neighbor nodes importance assignment.
The first component identifies the noise information in each hop and uses random walks to efficiently capture non-redundant graph structure information. The second component assigns importance weights to each neighbor nodes to help random walks capture more important information.

\textbf{Noise Information Identification.}
Considering the influence of noise, high hops often contain redundant information from low hops. 
We need to traverse the entire graph to identify the noise information of each hop.
Here, we use the de-noise matrix to record the noise information. Given the hop number $h$, the de-noise matrix of target node $v_i$ is defined as $M_{i}^h$ :
\begin{equation}
\begin{aligned}
&M_{i}^h=\bigcup_{j=1}^{{N}_h} m_{ij},&&  m_{ij}=\left\{
\begin{aligned}
&1 & {\rm if}~distance(v_i, v_j)=h \\
&0 & {\rm elif}~distance(v_i, v_j)<h
\end{aligned}
\right.
\end{aligned}
\label{noise-level}
\end{equation}
where ${N}_h$ is the number of neighbor nodes within a distance h from the target node $v_i$. The $distance$ between $v_i$ and $v_j$ is the shortest path length. If $distance(v_i, v_j)=h$, $m_{ij}^h $ is set to 1, otherwise it is identified as noise information and set to 0.
The de-noise matrix of the entire graph can be expressed as $M^h = \cup_{i \in N}M_i$, where $N$ is all the nodes in the entire graph.
The de-noise matrix enables us to extract the pure information at each hop while ensuring low smoothness level.

Afterward, for each hop, we use the random walk function (i.e., RW) to capture the graph structure information of the current hop, and the de-noise matrix is combined to extract useful information from each hop:
\begin{equation}
W^h=RW(\mathcal{G}, M^h, T)\\
\label{noise-level}
\end{equation}
where $T$ is the number of random walks. By controlling the number of random walks, we achieve a favorable balance between accuracy and efficiency, making it well-suited for large-scale graphs.

\begin{table}[t]
\caption{{Theoretical complexity for existing scalable GNNs. $n$, $m$, and $f$ are the number of nodes, edges, and feature dimensions, respectively. $r$ is the number of edges for each hop using our method, $R$ is the number of random walks, and $c$ is the number of threads. $L$ corresponds to the number of times we aggregate features, $K$ is the number of layers in MLP classifiers.}}
\setlength{\tabcolsep}{1.5mm}
\centering
\begin{tabular}{*{1}{c}|*{1}{c}|*{1}{c}|*{1}{c}}
\hline
Method&Pre-processing&Training&Inference\\
\hline
S$^2$GC&$\mathcal{O}(Lmf)$&$\mathcal{O}(nf^2)$&$\mathcal{O}(nf^2)$\\
SIGN&$\mathcal{O}(Lmf)$&$\mathcal{O}(Knf^2)$&$\mathcal{O}(Knf^2)$\\
GAMLP&$\mathcal{O}(Lmf)$&$\mathcal{O}(Knf^2)$&$\mathcal{O}(Knf^2)$\\
GBP&$\mathcal{O}(Lmf + L\frac{\sqrt{m lg n}}{\varepsilon})$&$\mathcal{O}(Knf^2)$&$\mathcal{O}(Knf^2)$\\
\hline
RMask&$\mathcal{O}(L\frac{(nR + rf)+}{c}\frac{m}{c\varepsilon})$&plug&plug\\
\hline
\end{tabular}
\label{tab:1}
\end{table}

\begin{table*}[t]
\caption{{Experiment results of node classification prediction tasks on six datasets.}}
\setlength{\tabcolsep}{3mm}
\centering
\begin{tabular}{c|*{3}{c}|*{3}{c}}%
\hline
\multirow{2}{*}{Methods}
&
\multicolumn{3}{c|}{Citation datasets}&\multicolumn{3}{c}{OGB datasets}\\
\cline{2-7}
&$\textbf{Cora}$&$\textbf{Citeseer}$&$\textbf{Pubmed}$&$\textbf{ogbn-arxiv}$&$\textbf{ogbn-products}$&$\textbf{ogbn-papers100M}$\\
\hline
SIGN&$82.1\pm0.3$&$72.4\pm0.8$&$79.5\pm0.5$&$71.9\pm0.1$&$78.2\pm0.3$&$64.3\pm0.1$\\
SIGN+RMask&$\pmb{84.3\pm0.6}$&$\pmb{73.6\pm0.6}$&$\pmb{80.3\pm0.7}$&$\pmb{72.8\pm0.3}$&$\pmb{81.1\pm0.3}$&$\pmb{65.3\pm0.3}$\\
\hline
S$^2$GC&$82.5\pm0.3$&$73.0\pm0.2$&$79.6\pm0.3$&$71.8\pm0.3$&$77.1\pm0.5$&$64.7\pm0.4$\\
S$^2$GC+RMask&$\pmb{83.8\pm0.5}$&$\pmb{73.8\pm0.5}$&$\pmb{80.2\pm0.5}$&$\pmb{72.7\pm0.3}$&$\pmb{78.6\pm0.8}$&$\pmb{65.5\pm0.6}$\\
\hline
GBP&$83.5\pm0.7$&$72.6\pm0.5$&$80.6\pm0.4$&$71.4\pm0.2$&$77.3\pm0.3$&$64.7\pm0.5$\\
GBP+RMask&$\pmb{84.3\pm0.6}$&$\pmb{73.7\pm0.5}$&$\pmb{81.1\pm0.6}$&$\pmb{71.6\pm0.6}$&$\pmb{78.4\pm0.4}$&$\pmb{65.5\pm0.6}$\\
\hline
GAMLP&$82.3\pm0.4$&$72.6\pm0.6$&$79.1\pm0.7$&$71.9\pm0.3$&$80.3\pm0.3$&$64.4\pm0.2$\\
GAMLP+RMask&$\pmb{83.6\pm0.4}$&$\pmb{73.3\pm0.5}$&$\pmb{80.3\pm0.5}$&$\pmb{72.9\pm0.2}$&$\pmb{81.4\pm0.4}$&$\pmb{65.2\pm0.4}$\\
\hline
\end{tabular}
\label{tab:4}
\end{table*}

\textbf{Neighbor Nodes Importance Assignment.}
Compared to existing model-simplification-based methods, the masking mechanism provides deeper and high-efficiency advantages, as it avoids the noise information by inter-hop parallel extraction of pure information from the graph structure.
However, due to the uniform sampling-based random walk method, this approach assigns equal importance to all neighbor vertices, which may not be expressive enough to capture the most important nodes from the graph.
To overcome this problem, we adopt a biased random walk based on neighbor node importance.
Specifically, we use Personalized PageRank~\cite{prfast} to get neighbor node importance. Because it can calculate the correlation of all neighbor nodes concerning the target nodes and can be efficiently computed using the approximation techniques to facilitate the scalability for large-scale graphs:
\begin{equation}
\begin{aligned}
\mathbf{S}= \alpha(\mathbf{I}-(1-\alpha)\mathbf{\hat{A}})^{-1}.
\end{aligned}
\label{eq-local-multi-2}
\end{equation}
where $\mathbf{I}$ is the identity matrix, $\alpha \in (0, 1]$  is the random walk restart probability, $\mathbf{\hat{A}}$ is the normalized matrix of $\mathbf{A}$.
$\mathbf{S}$ is the importance score matrix for each node.
We then use $\mathbf{S}$ in Eq. (3) to guide the random walks:
\begin{equation}
\begin{aligned}
W^h=RW(\mathcal{G}, M^h, T, \mathbf{S})\\
\end{aligned}
\label{eq-local-multi-2}
\end{equation}
by assigning the importance weight to each edge in the graph, the direction of the random walks can be guided so that it can capture more important de-noise information. Note that neighbor nodes importance assignment is optional and only needs to be performed once in the pre-processing stage to further improve accuracy.
Algorithm 1 outlines the overall process of RMask.

\textbf{Time Analysis.}
Table~\ref{tab:1} compares the time complexity of RMask with several representative sampling GNNs and model-simplification GNNs.
For model-simplification GNNs, in the pre-processing step, the time complexity of most linear models is $\mathcal{O}(Lmf)$, GBP conducts this process approximately with a bound of $\mathcal{O}(Lmf + L \frac{\sqrt{m lg n}}{\varepsilon})$, where $\varepsilon$ is an error threshold.
The time complexity of the serial version for RMask is $\mathcal{O}(L(nR + rf)+\frac{m}{\varepsilon})$. By running RMask in parallel using $c$ threads, the time complexity of the parallel version for RMask is $\mathcal{O}(L\frac{(nR + rf)}{c}+\frac{m}{c\varepsilon})$. Compared with model-simplification GNNs, the time complexity of our method is significantly lower.
As a plug-and-play module for model-simplification GNNs, RMask inherits the training model, ensuring consistent time and memory complexity during the training phase.

\section{EXPERIMENTS}
In this section, we execute comprehensive experiments to evaluate the proposed RMask for the node classification task on six widely-used graphs including Cora, Citeseer, Pubmed~\cite{GCN}, ogbn-arxiv (arxiv), ogbn-products (products), and ogbn-papers100M (papers100M)~\cite{ogb}. 
The details about all experiment settings and the network configurations are reported in the Appendix.
We showcase the benefits of RMask through five distinct perspectives: (1) a comparison from end-to-end with state-of-the-art model-simplification methods, (2) analyzing the ability towards deeper architecture, (3) analyzing the trade-off between efficiency and accuracy, (4) analyzing efficiency, (5) ablation study.

\begin{figure}[h]
    \centering
    \begin{subfigure}{0.49\linewidth}
        \includegraphics[width=\linewidth]{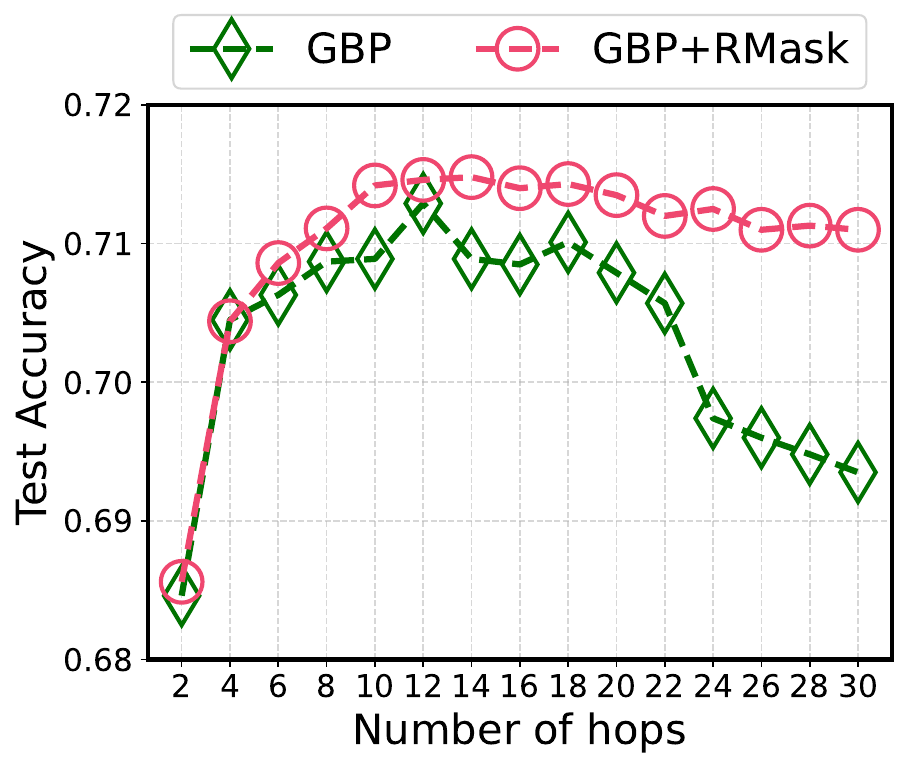}
        \caption{GBP + RMask}
    \end{subfigure}
    \begin{subfigure}{0.49\linewidth}
        \includegraphics[width=1.03\linewidth]{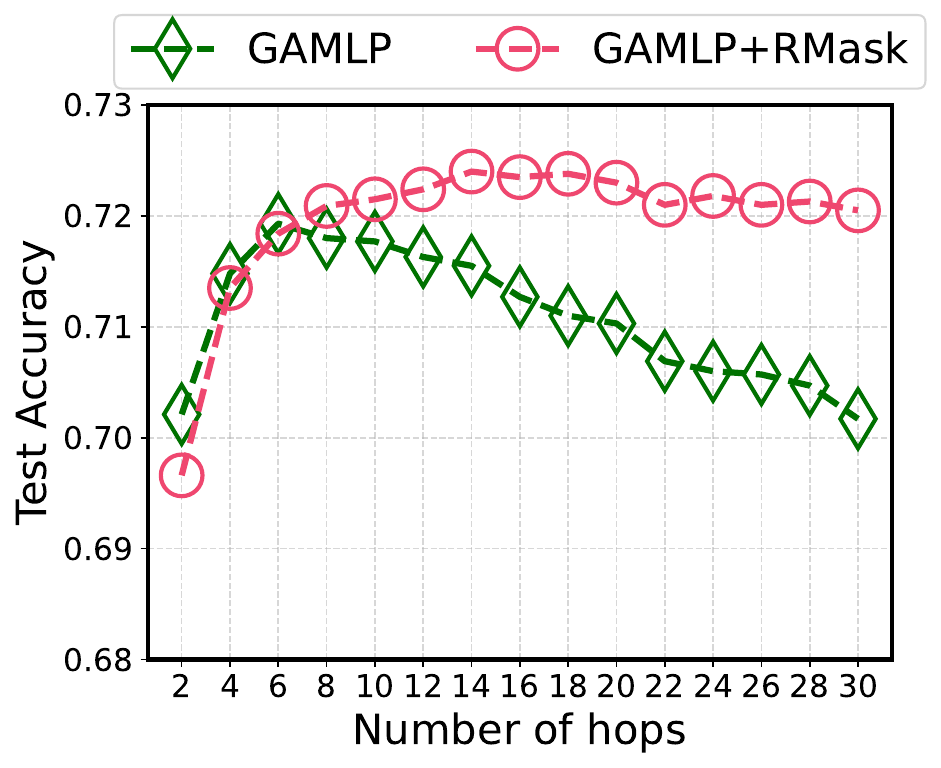}
        \caption{GAMLP + RMask}
    \end{subfigure}
\caption{Test accuracy on ogbn-arxiv along the propagation steps with two model-simplification GNNs.}
\label{fig:6}
\end{figure}

\begin{figure*}[t]
    \centering
    \begin{subfigure}{0.33\linewidth}
        \includegraphics[width=\linewidth]{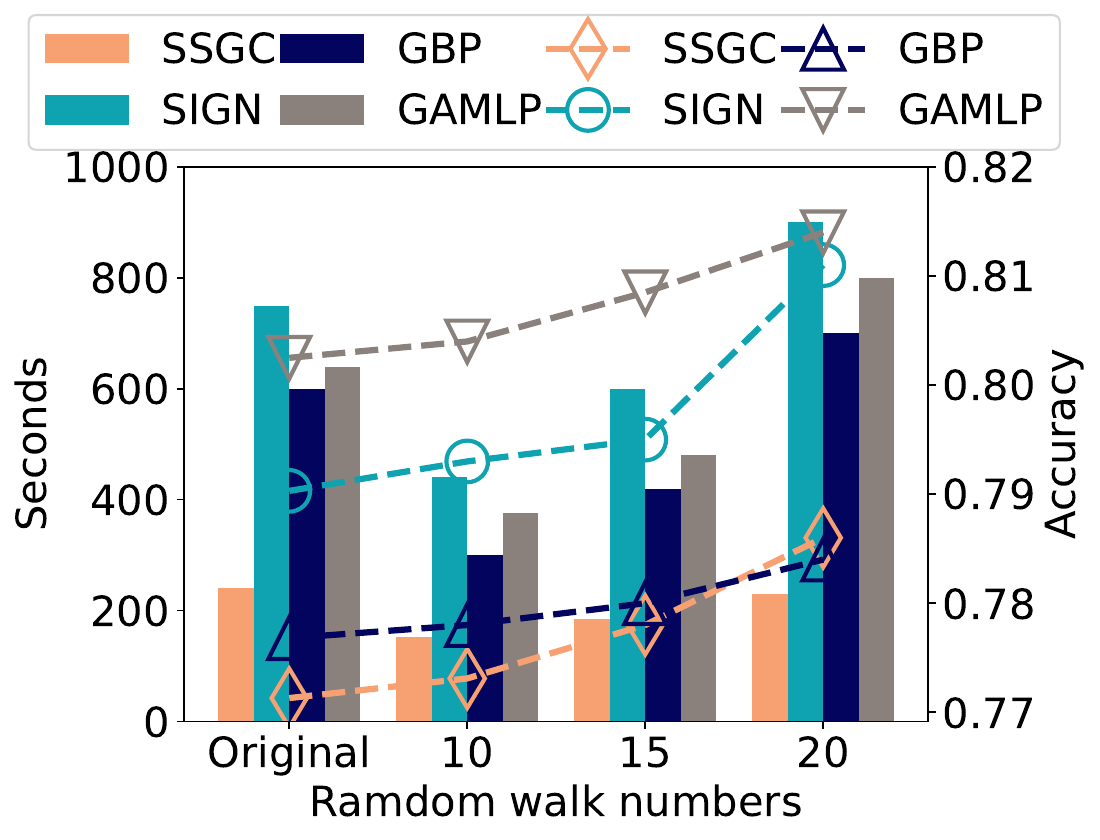}
        \caption{Model-simplification GNNs + RMask}
    \end{subfigure}
    \begin{subfigure}{0.33\linewidth}
        \includegraphics[width=\linewidth]{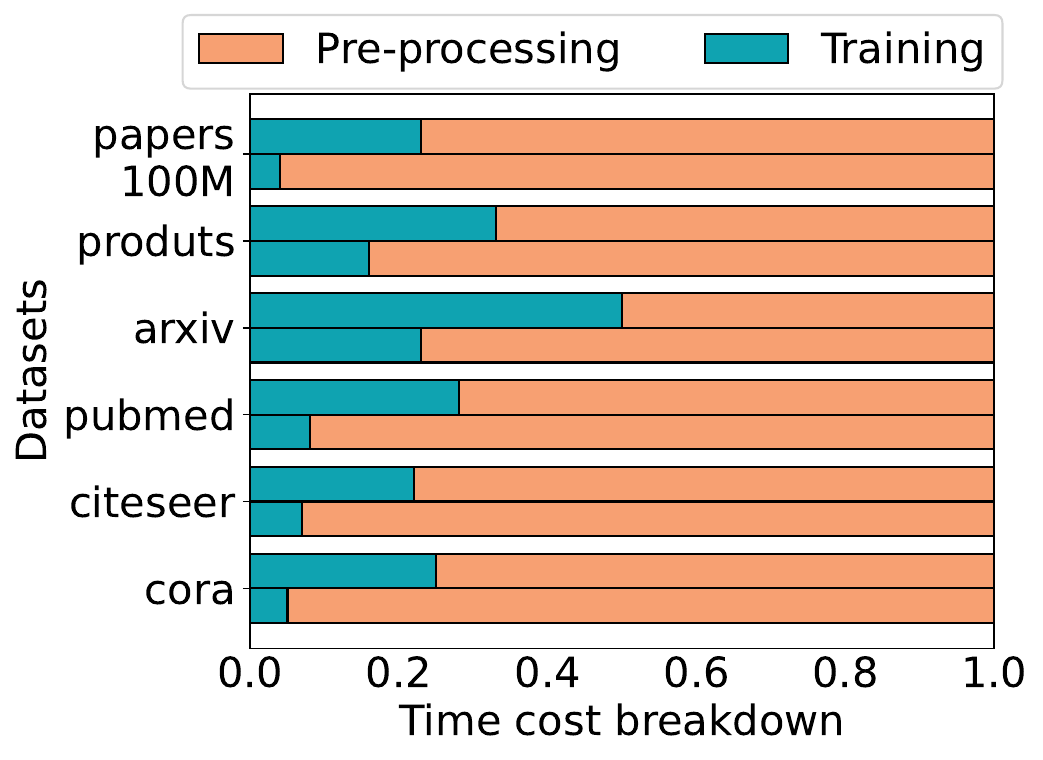}
        \caption{S$^2$GC vs. S$^2$GC+RMask}
    \end{subfigure}
    \begin{subfigure}{0.33\linewidth}
        \includegraphics[width=\linewidth]{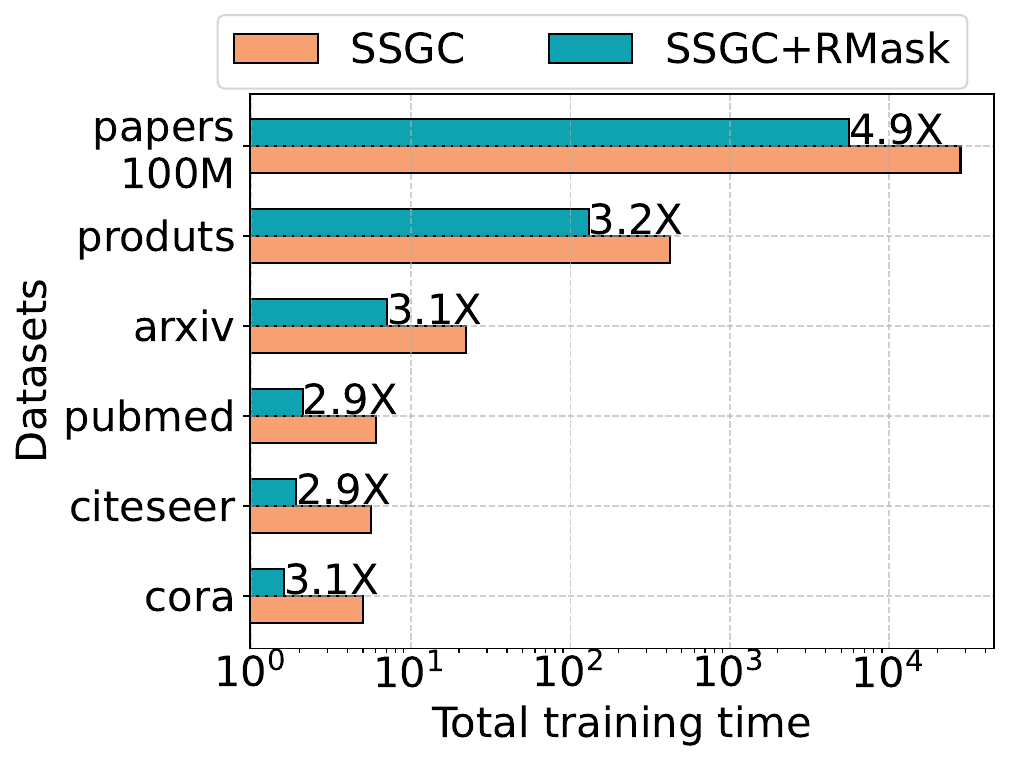}
        \caption{S$^2$GC vs. S$^2$GC+RMask}
    \end{subfigure}
    \caption{(a) Trade-off between efficiency and accuracy on ogbn-products. (b) Time cost breakdown. (c) Speedup analysis.}
\label{fig:5}
\end{figure*}

\begin{table*}[t]
\caption{{Ablation study to verify the effectiveness of RMask.}}
\centering
\begin{tabular}{{c}*{4}{c}|{c}*{3}{c}}
\hline
&\multicolumn{3}{c}{Cora}& &\multicolumn{3}{c}{Pubmed}\\
\hline
&Original&Variant\#1&Variant\#2&Variant\#3&Original&Variant\#1&Variant\#2&Variant\#3\\
\hline
r = 10&$82.8\pm0.6$&$82.1\pm0.3$&$82.3\pm0.2$&
$83.3\pm0.6$&
$79.3\pm0.7$&$79.5\pm0.5$&$78.8\pm0.4$&
$79.9\pm0.8$\\
r = 15&$84.3\pm0.6$&$82.1\pm0.3$&$83.9\pm0.2$&
$84.7\pm0.4$&$
80.3\pm0.7$&$79.5\pm0.5$&$79.8\pm0.5$&
$80.8\pm0.1$\\
r = 20&$84.6\pm0.5$&$82.1\pm0.3$&$84.1\pm0.3$&
$84.9\pm0.2$&
$80.8\pm0.6$&$79.5\pm0.5$&$80.2\pm0.4$&
$81.4\pm0.2$\\
\hline
\end{tabular}
\label{exp6}
\end{table*}

\subsection{End-to-end Comparison}
In Table~\ref{tab:4}, we present the results of node classification prediction using SIGN, S$^2$GC, GBP, and GAMLP on six datasets, both with and without our method. 
The experimental results demonstrate that when equipped with RMask, SIGN, S$^2$GC, GBP, and GAMLP all achieve superior performance compared to their respective original versions across all six datasets. 
Notably, since each baseline method represents a different combination approach for propagation steps, our proposed RMask can seamlessly integrate with all model-simplification GNNs, delivering enhanced performance without sacrificing scalability.

\subsection{Towards Deeper Architecture}
Existing model-simplification GNNs often suffer from over-smoothing issues when the propagation depth is large, leading to indistinguishable node representations and poor predictive performance. 
In this subsection, we perform experiments on the ogbn-arxiv dataset to demonstrate that model-simplification GNNs, when equipped with RMask, can effectively handle large $\textbf{P}$ operations. 
Figure~\ref{fig:6} depicts the results as we increase the number of $\textbf{P}$ operations from 1 to 30, with the green line indicating the original version and the red line representing the version equipped with RMask. 
While GBP and GAMLP experience a significant decline in performance as the number of $\textbf{P}$ operations increases, the predictive accuracy of GBP + RMask, and GAMLP + RMask either remain stable or only exhibit slight decreases. 
This stark contrast highlights the effectiveness of RMask in mitigating over-smoothing problem. Thus, utilizing RMask, model-simplification methods can leverage deep information more effectively, resulting in higher accuracy.

\subsection{The Trade-off between Efficiency and Accuracy}
\label{sec5:E}
Considering the high computational overhead incurred when processing large-scale graphs. We make a trade-off between efficiency and accuracy by controlling the number of random walks. As shown in Figure~\ref{fig:5}(a), we inserted RMask into four model-simplification GNNs on the ogbn-products dataset, the number of random walks are 10, 15, and 20, and compared with the original version (Original). The barplots represent the runtime and the lines represent the accuracy.
For all methods integrated with RMask, the accuracy is higher than the original version when random walk numbers are larger than 10, and the accuracy can be further improved as the number of random walks increases. 
In addition, high efficiency is guaranteed by controlling the number of random walks.
The comparative outcomes highlight that employing RMask as a plugin module yields satisfactory results with a minimal number of random walks.

\subsection{Efficiency Analysis}
To verify the efficiency of RMask, we first conducted a time cost breakdown of the S$^2$GC model on six datasets with five random walks. In Figure~\ref{fig:5}(b), the bottom bar of each dataset represents the original version, while the top bar represents the version equipped with RMask. Our method successfully reduces the proportion of pre-processing overhead in end-to-end training, particularly for large-scale graph datasets. 
In Figure~\ref{fig:5}(c), we further compare the end-to-end training cost. In all the datasets, S$^2$GC + RMask exhibits more than 2.9 times faster compared to S$^2$GC. And the speedup is even greater on large-scale data sets, up to 4.9 times. This outcome shows the efficiency of our approach.

\subsection{Ablation Study}
To conduct a comprehensive study of the proposed RMask, we performed ablation studies into two RMask variants. The original is the RMask with the noise masking mechanism, we use S$^2$GC+ RMask as the original version. Variant\#1 does not use noise information identification. Variant\#2 does not use the neighbor nodes importance assignment. Variant\#3 utilizes another over-smoothing solution, DAGNN~\cite{DAGNN}, as an orthogonal method for model training.

We conduct experiments with different numbers of random walks $r$ on the above three variants on the Cora and Pubmed datasets, respectively. r is set to 10, 15, 20.
Two observations can be made from Table~\ref{exp6}.

Firstly, Variant\#1 and Variant\#2 have different contributions to the performance. The noise information identification component can help RMask overcome the over-smoothing problem and make model-simplification GNNs go deeper. As the number of hops increases, the accuracy will not decrease. The neighbor nodes importance assignment component can help RMask further improve performance.
Secondly, from variant\#3, we can see that RMask can be orthogonal to other over-smoothing methods, thus further improving the accuracy.

\section{Conclusion}
This paper introduces RMask, a plug-and-play module designed to enhance existing model-simplification GNNs in exploring deeper graph structures at higher speeds. 
Unlike existing model-simplification GNNs focus on improving the combination method of propagated features.
RMask offers a novel perspective by enhancing the utilization of valuable information at each hop.
In the pre-processing step, RMask employs a mask method to eliminate noise information at each hop. To reduce the high overhead of pre-processing, RMask employs random walks to achieve a good trade-off between efficiency and accuracy.
As a plug-in method,re RMask seamlessly integrates with most model-simplification GNNs.
Experimental results on six real-world datasets demonstrate that RMask effectively enhances the accuracy and efficiency of model-simplification GNNs.

\section*{ACKNOWLEDGMENT}
This work is supported by National Natural Science Foundation of China (U22B2037, 92470121, 62402016), research grant No. IPT-2024JK29, the Fund of Kunpeng and Ascend Center of Excellence (Peking University), and High-performance Computing Platform of Peking University.

\bibliography{HGNNbibtex.bib}

\appendix

\section{Appendix}

\subsection{More Related Works}
\textbf{Sampling GNNs.}
An intuitive scalable approach is to use sampling techniques. 
Existing sampling work is usually divided into three categories: node level, layer level, and graph level.
As a node-level sampling method, GraphSAGE~\cite{GraphSAGE} samples the target nodes as a mini-batch and samples a fixed-size set of neighbors for computing. VR-GCN~\cite{VR-GCN} analyzes the variance reduction on node-wise sampling, and it can reduce the size of samples with an additional memory cost. In the layer level, Fast-GCN ~\cite{FastGCN} samples a fixed number of nodes at each layer, and ASGCN~\cite{ASGCN} proposes adaptive layer-wise sampling with better variance control. For the graph level sampling, Cluster-GCN ~\cite{ClusterGCN} clusters the nodes and only samples the nodes in the clusters, and GraphSAINT~\cite{GraphSAINT} directly samples a subgraph for mini-batch training.

\textbf{Model-simplification GNNs.}
In addition to sampling GNNs, another scalable approach is model-simplification GNNs. Based on Eq. (1), several model-simplification works have been proposed~\cite{SGC, S2GC, GBP} for scalable GNNs, which combine features at a finer granularity, i.e., hop-wise. For example, SIGN~\cite{SIGN} concatenates neighbor-aggregated features from different propagation layers: $[\mathbf{X}^{(0)}\mathbf{W}_0,\cdots, \mathbf{X}^{(k)}\mathbf{W}_k]$, while S$^2$GC~\cite{S2GC} proposes a simple spectral graph convolution to average the propagated features in different propagation layers: $\mathbf{X}^{(k)} = \sum_{i=0}^{k}\mathbf{\hat A}^i\mathbf{X}^{(0)}$. 
GBP~\cite{GBP} further improves the combination process by weighted averaging as $\mathbf{X}^{(k)} = \sum_{i=0}^{k}w_l\mathbf{\hat A}^i\mathbf{X}^{(0)}$ with the layer weight $w_l = \beta(1-\beta)^l$.
GAMLP~\cite{GAMLP} further considers feature propagation from a node-wise perspective, with each node having a personalized combination of the different layers of propagated features.

\emph{Comparison.}
Unfortunately, despite the focus of these models on combining different hops and improving the model design, they still face over-smoothing problem when the $\textbf{P}$ operation deepens. RMask addresses this problem by eliminating the noise problem generated within each hop, which hinders the effective 

\begin{table*}[t]
\renewcommand\arraystretch{1.4}
\caption{\textbf{Overview of datasets.}}
\centering
\begin{tabular}{*{7}{c}}%
\toprule%
$\textbf{Dataset}$&$\textbf{\#Nodes}$&$\textbf{\#Features}$&$\textbf{\#Edges}$&$\textbf{\#Classes}$&$\textbf{\#Train/Val/Test}$&$\textbf{Description}$\\
\midrule%
Cora&2,708&1,433&5,429&7&140/500/1000&citation network\\
Citeseer&3,327&3,703&4,732&6&120/500/1000&citation network\\
Pubmed&19,717&500&44,338&3&60/500/1000&citation network\\
\midrule%
ogbn-arxiv&169,343&128&1,166,243&40&91K/30K/49K&citation network\\
ogbn-products&2,449,029&100&61,859,140&47&196K/49K/2,204K&co-purchasing network\\
ogbn-papers100M&111,059,956&128&1,615,685,872&172&1200k/200k/146k&citation network\\
\bottomrule
\end{tabular}
\label{tab:3}
\end{table*}

\begin{figure*}[t]
    \centering
    \begin{subfigure}{0.32\linewidth}
        \includegraphics[width=\linewidth]{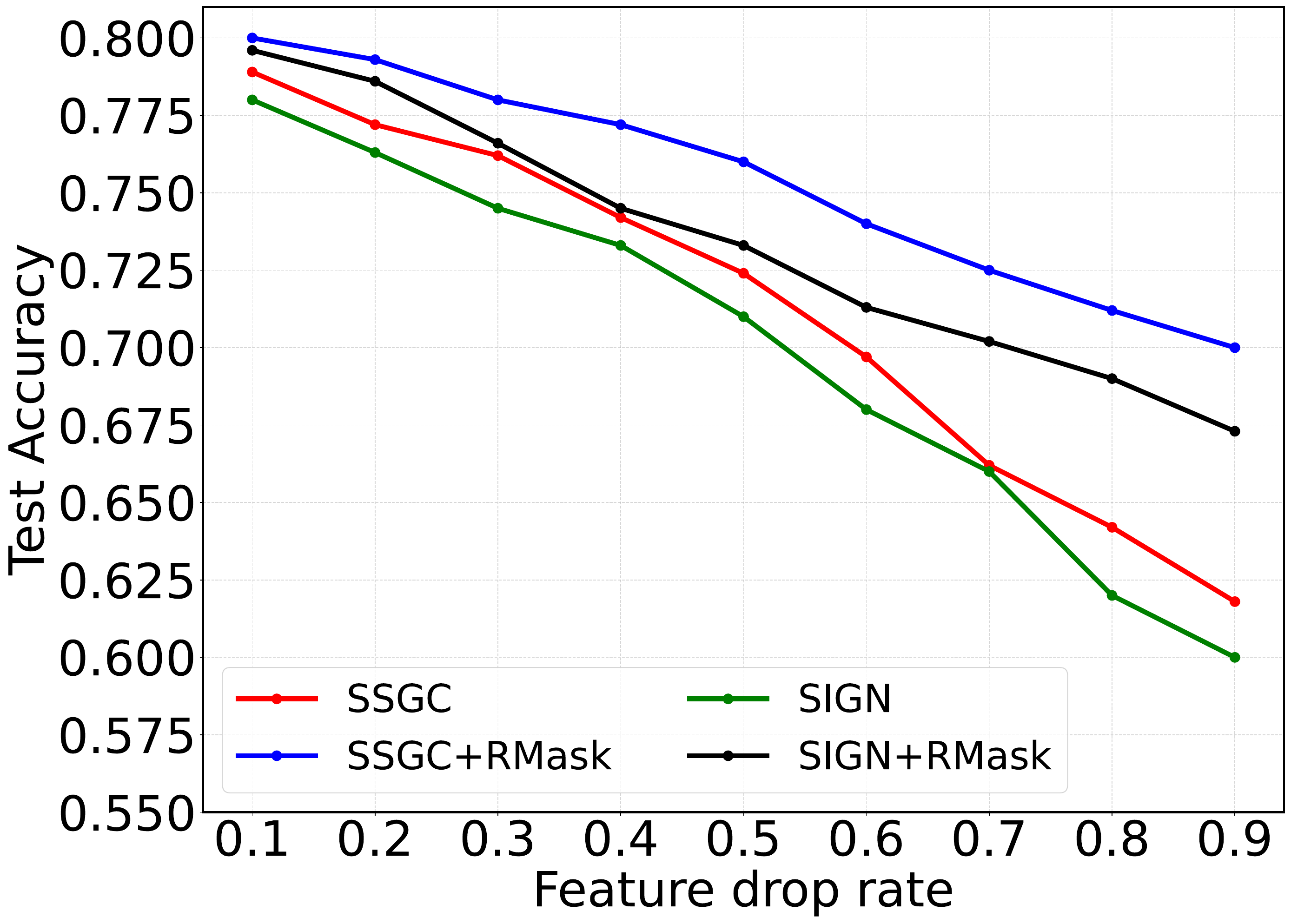}
        \caption{Feature sparsity}
    \end{subfigure}
    \begin{subfigure}{0.32\linewidth}
        \includegraphics[width=\linewidth]{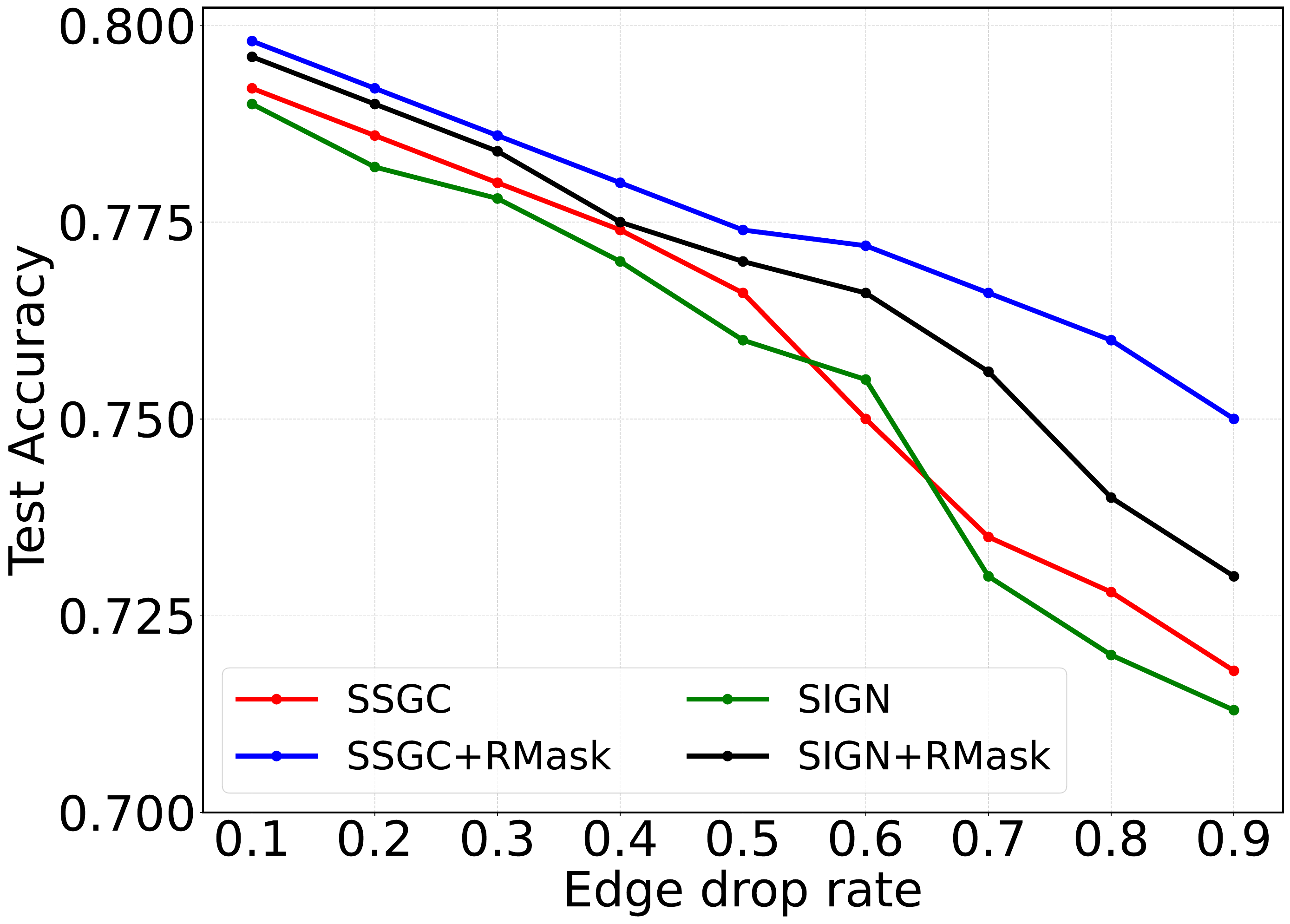}
        \caption{Edge sparsity}
    \end{subfigure}
    \begin{subfigure}{0.32\linewidth}
        \includegraphics[width=\linewidth]{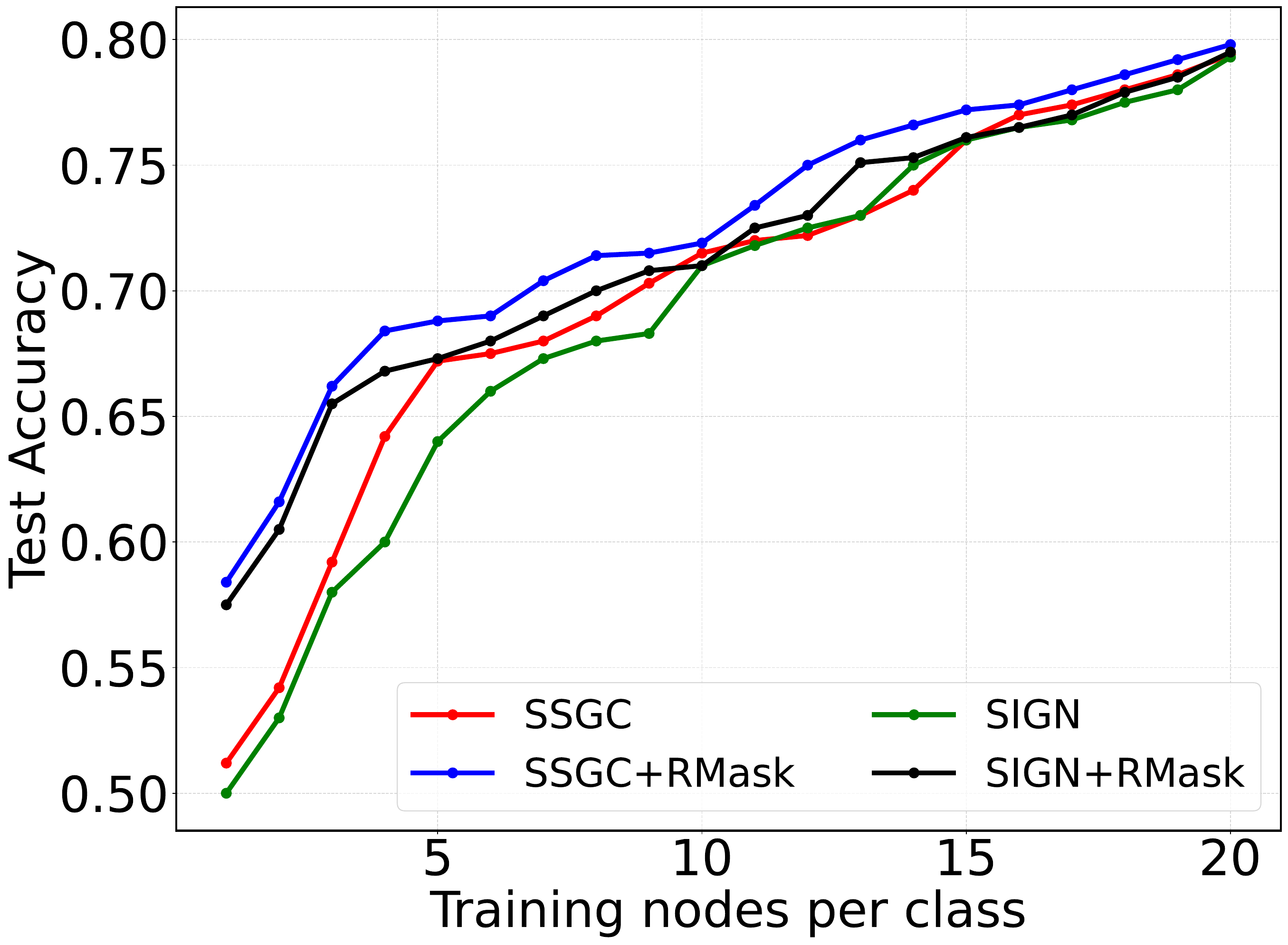}
        \caption{Label Sparsity}
    \end{subfigure}
    \caption{Test accuracy on the Pubmed dataset under different levels of feature, edge and label sparsity.}
\label{fig:9}
\end{figure*}
\subsection{More Experimental Details}
\textbf{Datasets.} We adopt three popular citation network datasets (Cora, Citeseer, PubMed)~\cite{GCN} and three large-scale OGB datasets (ogbn-arxiv, ogbn-products, ogbn-papers100M)~\cite{ogb} to evaluate the predictive accuracy of each method on the node classification task.

{For three popular citation network datasets, papers from different topics are considered nodes, and the edges are citations among the papers. The node attributes are binary word vectors, and class labels are the topics the papers belong to.
For three large-scale OGB datasets, ogbn-arxiv is a directed graph, representing the citation network among all Computer Science (CS) arXiv papers indexed by MAG. ogbn-products is an undirected and unweighted graph, representing an Amazon product co-purchasing network. ogbn-papers100M is a directed citation graph of 111 million papers indexed by MAG.}

Table~\ref{tab:3} presents an overview of these six datasets. For all datasets, we adopt the official training/validation/test split.

\textbf{Baseline Methods.} To evaluate the performances of our RMask, we integrate the proposed module into four model-simplification GNNs, namely SIGN~\cite{SIGN}, S$^2$GC~\cite{S2GC}, GBP~\cite{GBP}, and GAMLP~\cite{GAMLP}. 
We then compare the performance of these modified versions with their original versions. To alleviate the influence of randomness, we repeat each method ten times.

\textbf{Hyperparameters.} 
The hyperparameters in four model-simplification GNNs are set according to the original paper. For S$^2$GC, GBP, GAMLP, and SIGN equipped with RMask, the hidden size is set to 64, 64, 128, and 512 in small datasets Cora, Citeseer, and Pubmed respectively. In large datasets ogbn-arxiv, ogbn-products and ogbn-papers100M, the hidden size is set to 512.
As for the dropout percentage and the learning rate, we use a grid search from \{0, 0.1, 0.2, 0.3, 0.4, 0.5\} and \{0.1, 0.01, 0.001\} respectively.
For the propagation steps and a number of random walks, we use a grid search from \{6 - 15\} and \{5, 10, 15, 20, 25, 30\} respectively. We train our models using Adam optimizer\cite{Adam} during training.
For the training budget, we train every small-scale dataset with 300 epochs and we terminate the training process if the validation accuracy does not improve for 100 consecutive steps. For large-scale datasets, we train with 1500 epochs and terminate for 200 consecutive steps.

\textbf{Experiment Environment.}
The experiments are conducted on a machine with Intel(R) Xeon(R) Gold 5120 CPU @ 2.20GHz and a single TITAN RTX GPU with 24GB GPU memory. The operating system of the machine is Ubuntu 16.04. We use Python 3.7, Pytorch 1.12.1, and CUDA 10.1.

\subsection{The Influence of Graph Sparsity on RMask}
Given that RMask can assist model-simplification GNNs in exploring deeper propagation steps, we conducted simulations in extremely sparse scenarios in the real world. Additionally, we designed three sparsity settings on the Pubmed dataset to evaluate the performance of our proposed RMask when encountering edge sparsity, label sparsity, and feature sparsity issues, respectively. We choose the representative methods, S$^2$GC and SIGN, which correspond to the two most commonly used combinations of model-simplification GNNs (weighted summation and concatenate) as our baseline. We then incorporate RMask into these methods to evaluate their performance.

\textbf{Feature Sparsity.}
In real-world scenarios, it is common for some nodes in the graph to have missing features. To simulate this, we randomly mask a portion of node features, with the masking rate varying from 0.1 to 0.9. The results depicted in Figure~\ref{fig:9}(a) demonstrate that our proposed RMask significantly enhances the anti-interference capabilities of the two baseline methods when confronted with feature sparsity.
The predictive performance of the S$^2$GC + RMask and SIGN + RMask can drop slower when the drop ratio gets larger and the number of training nodes gets fewer, demonstrating that RMask can help model-simplification GNNs perform better by exploring more deeper and useful graph information.

\textbf{Edge Sparsity.}
We randomly drop some edges from the original graph to simulate edge sparsity. The removed edges are the same across all the compared methods, with a fixed edge remaining rate ranging from 0.1 to 0.9. The experimental results in Figure~\ref{fig:9}(b) demonstrate that all the compared methods exhibit similar performance, as edges play a vital role in GNN methods. However, it is evident from the results that both S$^2$GC and SIGN show significant performance improvement when equipped with RMask.

\textbf{Label Sparsity.}
In the label sparsity setting, we examine the impact of varying the number of training nodes per class from 1 to 20 on the test accuracy of each compared method. 
Given the limited supervision signal, the classifier may not be effectively trained, emphasizing the need for improved node embedding in the model-simplification propagation process.
The experimental results depicted in Figure~\ref{fig:9}(c) demonstrate a consistent increase in the test accuracies of all compared methods as the number of training nodes per class increases. Notably, throughout the experiment, both S$^2$GC and SIGN, equipped with RMask, consistently outperform their original versions.

The aforementioned evaluation across three distinct levels of sparsity demonstrates the significant assistance provided by RMask to model-simplification GNNs. When applied to sparse graphs, model-simplification GNNs require additional \textbf{P} operations due to the presence of hidden information that can potentially be accessed through long-range connections.
Therefore, the enhanced capability of deep exploration brought by RMask plays a pivotal role in significantly improving the performance of model-simplification GNNs on sparse graphs.

\end{document}